\documentclass[10pt,journal]{IEEEtran}
\ifCLASSOPTIONcompsoc
\else
\fi

\ifCLASSINFOpdf
  \usepackage[pdftex]{graphicx}
  \graphicspath{{../pdf/}{../jpeg/}}
  \DeclareGraphicsExtensions{.pdf,.jpeg,.png}
\else
  \usepackage[dvips]{graphicx}
  \graphicspath{{../eps/}}
  \DeclareGraphicsExtensions{.eps}
\fi

\usepackage{algorithm}
\usepackage{algpseudocode}
\usepackage{amsmath}
\usepackage{graphics}
\usepackage{epsfig}
\usepackage{times}
\usepackage{graphicx}
\usepackage{amssymb}
\usepackage{amsfonts}
\usepackage{color}
\usepackage{bm}
\usepackage{multirow}
\usepackage{algpseudocode}
\usepackage{graphics}
\usepackage{threeparttable}
\usepackage{color}
\usepackage[normalem]{ulem}
\usepackage{multirow}
\usepackage{float}
\usepackage{amsfonts}
\usepackage{bm}
\usepackage{array}
\usepackage[table]{xcolor}
\DeclareMathOperator*{\argmin}{argmin}
\newcommand{\M}{\overline{M}^t}

\begin{document}
%
\title{Selective Video Object Cutout}
\author{Wenguan Wang, Jianbing Shen,~\IEEEmembership{Senior Member,~IEEE},
and Fatih Porikli,~\IEEEmembership{Fellow,~IEEE}
\IEEEcompsocitemizethanks{%
\IEEEcompsocthanksitem
This work was supported in part by the National Basic Research Program of China (973 Program) (No. 2013CB328805),
the Australian Research Council's Discovery Projects funding scheme (DP150104645), and the Fok Ying-Tong Education Foundation for Young Teachers.
Specialized Fund for Joint Building Program of Beijing Municipal Education Commission. (Corresponding author: Jianbing Shen)
\IEEEcompsocthanksitem W. Wang, J. Shen are with Beijing Laboratory of Intelligent Information Technology,
School of Computer Science, Beijing Institute of Technology, Beijing 100081, P. R. China.
(Email: {shenjianbing@bit.edu.cn})
\IEEEcompsocthanksitem F. Porikli is with the Research School of Engineering, Australian National University, Australia.
(Email: {fatih.porikli@anu.edu.au})
}
\thanks{}
}
\markboth{IEEE Transactions on Image Processing} 
{Shell \MakeLowercase{\textit{et al.}}: Bare Demo of IEEEtran.cls
for Computer Society Journals}

\maketitle
\begin{abstract}
Conventional video segmentation approaches rely heavily on appearance models. Such methods often use appearance descriptors that have limited discriminative power under complex scenarios. To improve the segmentation performance, this paper presents a pyramid histogram based confidence map that incorporates structure information into appearance statistics. It also combines geodesic distance based dynamic models. Then, it employs an efficient measure of uncertainty propagation using local classifiers to determine the image regions where the object labels might be ambiguous. The final foreground cutout is obtained by refining on the uncertain regions. Additionally, to reduce manual labeling, our method determines the frames to be labeled by the human operator in a principled manner, which further boosts the segmentation performance and minimizes the labeling effort. Our extensive experimental analyses on two big benchmarks demonstrate that our solution achieves superior performance, favorable computational efficiency, and reduced manual labeling in comparison to the state-of-the-art.
\end{abstract}

\begin{IEEEkeywords}
Video cutout, video object segmentation, pyramid appearance model, local uncertainty propagation.
\end{IEEEkeywords}

%

\IEEEpeerreviewmaketitle

\section{Introduction}
\label{sec:introduction}
Video cutout aims at pixel-level labeling of video frames given the user annotations by taking advantage of the continuity of the video content. With the initialization of the human interactions on few frames (first frame or several key frames), the cutout method labels the rest frames via propagating the user annotations. This task contributes to a variety of computer vision applications including human action recognition, object detection and recognition, and video editing, to name a few.

In this context, many video cutout techniques \cite{Bai2009, Wang2015cosegment, Zhong2012, Ramakanth2014,Wang2005} have been proposed to propagate user annotations, adopting optical flow and spatiotemporally connected Markov chains as basic principles. Nevertheless, there are many challenging factors in inter-frame propagation including unreliable optical flow estimation, changing motion patterns, motion blur, and cluttered backgrounds that adversely affect the performance. To tackle these challenges, appearance models such as Gaussian Mixture Models (GMMs) and color histograms are often incorporated. However, the discriminative capability of appearance models diminishes when object-background color models become ambiguous. Besides, most appearance models only account for higher level image features that cannot be explicitly tailored for a particular configuration of local receptive fields.

Another concern with video cutout is its labor intensive nature. Although the state-of-the-art methods~\cite{Bai2009, Wang2015cosegment, Zhong2012, Fan2015, shankar2015} partly reduced the time-consuming burden of manual labeling, still a considerable human interaction is required. For long, interactive video cutout approaches focused only on segmentation engines, ignoring how the interaction between the user and the algorithm should be formed for an optimal utilization of the human effort. For instance, existing methods propagate annotations from an arbitrarily selected frame (\textit{e.g.}, the first frame) yet there is no guarantee that arbitrarily selected frames (or even human-selected frames) could provide sufficient information for an optimal cutout.

In this paper, we propose a novel pyramid model empowered confidence maps and an adaptive segmentation framework that minimizes inter-frame propagation errors for boundary-accurate foreground segmentation and automatic selection of the key frames for manual labeling. The confidence map is a probability density function over the image, which could be treated  as \textit{classifier} that assigns each pixel a probability of being foreground. Inspired by the multi-level histograms in image classification \cite{Grauman2005, Lazebnik2006}, we incorporate a spatial pyramid color histogram model with a distance-classification based dynamic model, for considering locality information. This model provides efficient and reliable representations at several spatial resolution levels at the same time. Thus, it improves the discriminative power of the confidence maps compared to conventional classifiers that apply only at a single level. Our use of global and local models results in significantly improved performance over \cite{Wang2005,Li2005} that use either local or global cues. In addition, we explicitly formulate a \textit{propagation uncertainty} term to identify the pixels where the label is ambiguous. Within such regions, we leverage on the local model to fuse appearance features such as color and shape. The segmentation is finally achieved by fusing confidence maps in a principled manner.

To make the best use of human feedback, our method selects the key frames for human annotation. To this end, a substantial effort is spent for pixel-level predictions in \cite{Sudheendra2012}, which precludes it from practical use. We revise it on superpixel granularity. Our prediction model makes the propagation algorithm computationally more efficient by eliminating motion estimation. Our source code will be available online \footnote{\textit{http://github.com/shenjianbing/videocutout}}.

To summarize, the main contributions of this work are threefold:
\begin{itemize}
\item It proposes a new pyramid model to capture appearance and location information. This model is supplemented with a set of distance-classification based dynamic foreground models. Contriving them as confidence maps, our method provides significantly improved foreground detection performance (Sec. \ref{sec:globalclassifier}).
\item It introduces a local classifier based estimation of propagation-uncertainty for effective handling of regions where ambiguous labeling may occur (Sec. \ref{sec:localclassifier}).
\item It incorporates an annotation frame selection technique to automatically determine the frames for human annotation, significantly reducing the labeling effort while further improving the segmentation results (Sec. \ref{sec:intelligentannotationframeselection}).
\end{itemize}

The remainder of the article is organized as follows.
A summarization of related work is introduced in Sec. \ref{sec:relatedwork}.
Our video cutout approach is presented in detail Sec. \ref{sec:ourapproach}.
We then experimentally demonstrate its robustness, effectiveness, and efficiency in Sec. \ref{sec:results}.
Finally, we draw conclusions in Sec. \ref{sec:conclusion}.

\section{Related Work}
\label{sec:relatedwork}
Video segmentation aims at offering a binary labeling mask for each frame, thus separating foreground object from the background region of a video. Broadly speaking, video segmentation algorithms can be categorized as automatic (unsupervised), and supervised methods, according to the level of supervision required. In this section, we provide a brief overview of previous works in video object segmentation along these two major categorizes: unsupervised video segmentation and interactive video segmentation.

\subsection{Unsupervised Video Segmentation}
Given a video sequence, unsupervised video object segmentation aims at automatically separating each pixel into foreground or background.
Unsupervised algorithms do not require any manual annotation but usually rely on certain assumptions about the foreground. Appearance or motion based cues are
typically employed for inferring foreground object~\cite{Brox2010,chen2015,Xu2012,xiao2016track}. More specially, some techniques \cite{Brox2010,chen2015,Fragkiadaki12,Papazoglou2013} emphasize the importance of motion information. The segmentation is achieved via tracking optical flow among several frames \cite{Brox2010,Fragkiadaki12} and clustering trajectories for selecting the moving objects. Other methods \cite{faktor2014,Wang2015,Wang2017saliency,Wang2017coseg} employ various
saliency stimuli \cite{wang2015consistent,wang2016Correspondence} for extracting salient moving objects as foreground. Addition to utilizing low level features, several recent methods~\cite{Lee2011,Ma2011,zhang2013} consider object-level cues via exploring the notion of what a generic object looks like. They approach video segmentation as a task of selecting primary object regions in object proposal domain. Via employing generic object candidate generation methods \cite{Endres2010,Alexe2010}, a lot of object proposals are generated in every image/frame. Then, the foreground object is selected from those object candidates according to different principles and inference strategies. More recently, deep learning model is employed for video segmentation \cite{fragkiadaki2015learning}, which gives promising results.

\subsection{Supervised Video Segmentation}
Compared with unsupervised video segmentation methods, interactive video segmentation extracts foreground object from video sequences with the guidance of human interaction. Typical interactive methods propagate the given annotations to the entire video sequence, by tracking them using spatiotemporal Markov random fields based probabilistic models \cite{Jain2014,Price2009, Badrinarayanan2010,budvytis2011,Sudheendra2012}, or frame-matching based propagation \cite{Bai2009, Zhong2012, Fan2015, Tsai2016}, and employing various features such as color, shape, and motion.

In graphics community, many supervised video segmentation approaches \cite{Wang2005,Bai2009,Zhong2012,Fan2015} are also proposed, often with intensive human interactions. An early work, named as Video SnapCut~\cite{Bai2009}, incorporates a set of local classifiers using multiple local image features and integrates a propagation based shape model into color models. Zhong \textit{et al.} \cite{Zhong2012} introduces directional classifiers to handle temporal discontinuities while remaining robust to inseparable color statistics. More recently, \cite{Fan2015} obtains the segmentation at each frame by transferring the foreground mask using nearest-neighbor fields.

Tracking of segmentation results in video sequences has also been investigated by many works~\cite{tsai2010,Sudheendra2012, li2013, Ramakanth2014, Perazzi2015,Wen_2015_CVPR, maerki2016,Wang2017super} in computer vision research. The segmentation is obtained by either solving an optimization problem with patch seams across frames \cite{Ramakanth2014}, using fully connected graphs to model long-range spatiotemporal connections \cite{Perazzi2015}, operating on bilateral space \cite{maerki2016}, or adopting super-trajectory for capturing more rich information \cite{Wang2017super}. Such methods often require human annotations in the first frame, and then track them in the rest of the video. Recently, the work in \cite{Sudheendra2012} proposed an active frame selection method to select a subset of frames that together minimize the total mislabeling risk over the entire sequence. However, this strategy is computationally expensive. The propagation error prediction is done using a forward and backward pixel flow model, which is impractical for user-in-the-loop applications. In contrast, our model for predicting propagation error is based on superpixel-level matching, which is more efficient.

\section{Our Approach}
\label{sec:ourapproach}
Given an input video sequence, our segmentation process starts by automatically selecting a set of key frames for human labeling. It then propagates those labels to the rest of the video. Using the initial annotations, a static pyramid classifier is established and supplemented with a set of geodesic distance based dynamic foreground models. Empowered with discriminative and predictive capacity, a group of localized classifiers is applied to generate final segmentation results based on propagation uncertainty.

In~Sec. \ref{sec:globalclassifier}, we first introduce static and dynamic global models for confidence map estimation within and across video frames. Then, in Sec. \ref{sec:localclassifier}, we detail our localized models and propagation uncertainty estimation. Although our key frame selection is performed first in segmentation process, we state such technique at last (Sec. \ref{sec:intelligentannotationframeselection}), since the selection criterion is derived from our segmentation method.

\begin{figure}
  \centering
      \includegraphics[width=1.0\linewidth]{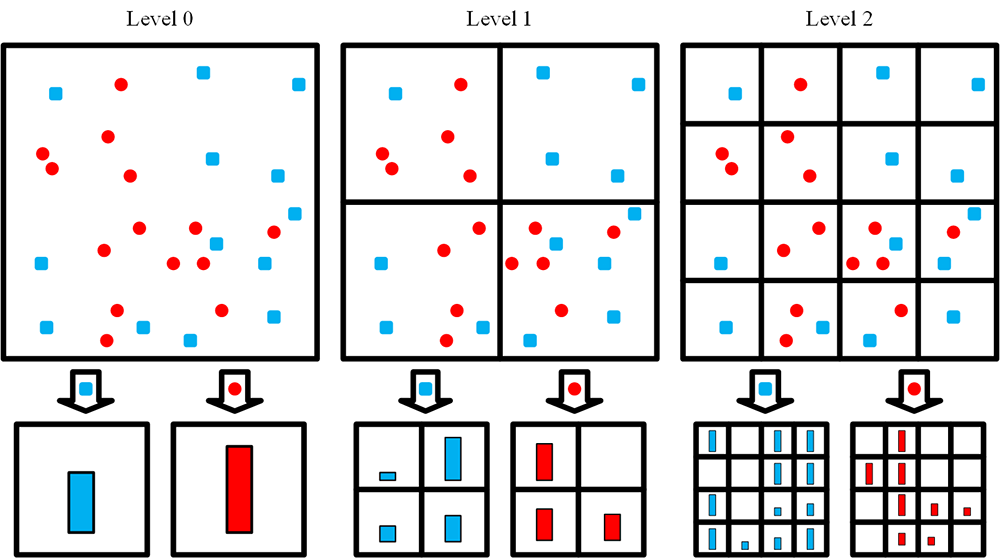}
\caption{Illustration of a three-level pyramid histogram model. The image has two types of color samples: foreground (red circles) and background (blue squares). We partition the image at three spatial resolution levels. For each level, we count the color samples in each bin, establishing a higher-dimensional histogram representation that accounts for color and location.}
\label{fig:PyramidModel}
\end{figure}

\subsection{Global Confidence Maps}
\label{sec:globalclassifier}
Given an input video sequence $\textbf{I} = \{I^{1},I^{2},{\cdots}, I^{N}\}$ containing $N$ frames and a subset of $K$ frames $\widehat{\textbf{I}} = \{I^{n_1},I^{n_2},{\cdots}, I^{n_K}\}$ with corresponding binary annotation masks $\widehat{\textbf{M}} = \{M^{n_1},M^{n_2},{\cdots}, M^{n_K}\}$ such that $\widehat{\textbf{I}} \subset \textbf{I}$, our goal is to assign each pixel a label $\textbf{L} = \{0, 1\}$ (\textit{background}\! =\! 0, \textit{foreground}\! =\! 1). We propagate initial annotations $\widehat{\textbf{M}}$ to the entire video. Each $M^{n_i}$ is a label matrix having the same dimensions as the image frame indexed by the 2D pixel coordinates $\textbf{x}$, where $M^{n_i}(\textbf{x}) \in \textbf{L}$. We begin by oversegmenting each frame by SLIC~\cite{Achanta2012} for computational efficiency. We then build global appearance models to assign each superpixel a probability of being foreground object. The number of superpixels is set to 2000 for a $1280\times720$ resolution image. We denote by $\textbf{Y}^t = \{Y^t_1,Y^t_{2},{\cdots}\}$ the superpixel set of frame $I^t$.

\subsubsection{Static Confidence Map}
Traditional global appearance models do not consider the spatial arrangements of features and fail to take the full advantage of the prior distributions available in user annotations. This motivated us to enhance histogram based models with the structural information by building a pyramid in the spatial domain then binning the feature space.

Our method repeatedly subdivides the frame into cells and computes histograms of color features in these cells. More specifically, it constructs a sequence of grids at resolutions $0, {\cdots}, L$ for each annotated frame $I^{n_k}$, such that the grid at level $\ell$ has $2^\ell$ cells along the coordinate directions. Then, for each grid $\kappa = 1, {\cdots}, 2^\ell$ in ceratin level $\ell$, it computes two color histograms, $H_F^{\ell, \kappa}$ and $H_B^{\ell, \kappa}$, from RGB color features at each level for all annotated frames $\widehat{\textbf{I}}$. Each pixel contributes into $H_F^{\ell, \kappa}$ and $H_B^{\ell, \kappa}$ according to its color value, label, and coordinate. Therefore, a global pyramid histogram model can be setup, consisting of a set of atomic histograms $\{H_F^{\ell, \kappa}\}_{\ell, \kappa}$ and $\{H_B^{\ell, \kappa}\}_{\ell, \kappa}$ at different grid levels. The foreground probability $\textsl{G}_{static}(Y^t_i)$ of superpixel $Y^t_i$ from the whole global appearance model is formulated as:
\begin{equation}
    \begin{aligned}
    \textsl{G}_{static}(Y^t_i)\!=\!\frac{1}{L}\sum\nolimits_{\ell=1}^{L}\sum\nolimits_{\kappa=1}^{2^\ell} \frac{H_F^{\ell, \kappa}(Y^t_i)}{H_F^{\ell, \kappa}(Y^t_i)+H_B^{\ell, \kappa}(Y^t_i)}.
    \end{aligned}
    \label{eq:1}
\end{equation}
Superpixel $Y^t_i$ is represented by its mean color and center position. If the center of $Y^t_i$ locates in the corresponding grid $\kappa$, $H^{\ell, \kappa}(Y^t_i)$ returns it histogram value according to its color; otherwise it returns 0. An example of constructing a three-level pyramid of an annotated frame is shown in Fig.~\ref{fig:PyramidModel}. Note the resolution of frame images is fixed, we vary the spatial resolution at which the color points are binned. This results in a higher-dimensional color histogram representation that carries more structure information and leads to more reliable classifications.

\begin{figure}[t]
  \centering
      \includegraphics[width=0.99 \linewidth]{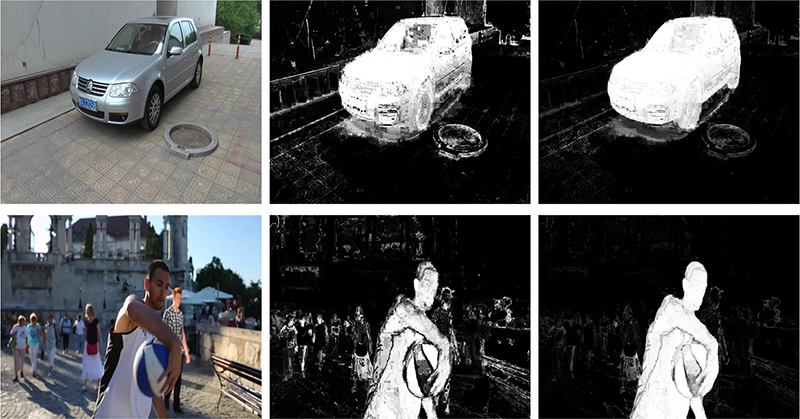}
           \\ \mbox{}\hfill (a) \hfill\mbox{}
     \mbox{}\hfill (b)  \hfill\mbox{}
     \mbox{}\hfill (c) \hfill\mbox{}
\caption{Comparison of our global pyramid appearance model to traditional color histograms. As visible, the foreground confidence maps generated by our global pyramid model (c) is more accurate than the ones generated by traditional color histograms (b).}
  \label{fig:PyramidVisual}
\end{figure}

In our implementation, each color channel is uniformly quantized into 32 bins, thus there is a total of $32^3$ bins in the histogram of each cell. The finest resolution $L$ is set as 3. We provide detailed discussion for $L$ in Sec.~\ref{sec:PPC}. Fig.~\ref{fig:PyramidVisual} shows two challenging sample frames where the foreground object and the background scene have overlapping color histograms, which confuses traditional classifiers built upon color statistics (Fig.~\ref{fig:PyramidVisual} (b)). In contrast, our pyramid appearance model fuses color and location information, resulting in a stronger discriminative power and more accurate probability maps (Fig.~\ref{fig:PyramidVisual} (c)).
\begin{figure*}
  \centering
      \includegraphics[width=1.0 \linewidth]{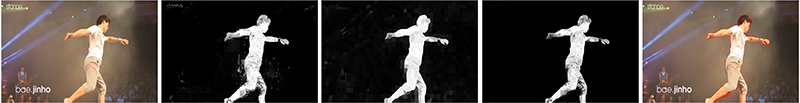}
  \\ \mbox{}\hfill (a) \hfill\mbox{}
     \mbox{}\hfill (b)  \hfill\mbox{}
     \mbox{}\hfill (c) \hfill\mbox{}
     \mbox{}\hfill (d) \hfill\mbox{}
     \mbox{}\hfill (e) \hfill\mbox{}
\caption{Illustration of our coarse segmentation. (a) Input frame $I^t$. (b) Foreground confidence map $G^t_{static}$ generated by the static global appearance model. (c) Foreground confidence map $G^t_{dynamic}$ generated by the dynamic global appearance model. (d) Combined foreground confidence map $G^t$ by consolidating (b) and (c). (e) Coarse segmentation results $\overline{M}^t$ via Eqn.~\ref{eq:7}.}
  \label{fig:globalvisual}
\end{figure*}
\subsubsection{Dynamic Confidence Map}
Our static pyramid histogram model offers far more valuable information than traditional appearance model, but it is also necessary to build a dynamic model for capturing inter-frame correspondence and accommodating dynamic variations between successive frames. Therefore, we develop a set of dynamic global classifiers, which estimate global confidence maps between consecutive frames and use geodesic distance, for complementing our static global appearance model.

For two adjacent frames $I^{t-1}$ and $I^t$, where frame $I^{t-1}$  with known segmentation $M^{t-1}$, we construct an undirected weighted graph $\mathcal{G} = \{\mathcal{V}, \mathcal{E}\}$, where $\mathcal{V} = \textbf{Y}^{t-1} \!\cup \textbf{Y}^{t}$. The edges $\mathcal{E}$ link spatially adjacent superpixels or temporally adjacent superpixels. The superpixels are spatially connected if they are adjacent in same frame. Temporally adjacent superpixels refer to the superpixels which belong to different frames but have overlap. Based on the graph structure, we derive a weight matrix $\mathcal{W}$. The $(a,b)$-th element of $\mathcal{W}$ indicates the weight of edge $e_{ab} \in \mathcal{E}$ between two connected nodes $v_a, v_b$:
 \begin{equation}
    \begin{aligned}
        w_{ab} = \| \mu(v_a) - \mu(v_b)\|,
    \end{aligned}
    \label{eq:2}
 \end{equation}
where $\mu(\cdot)$ represents the mean color value of the node (superpixel) and $v_a, v_b \in \mathcal{V}$.
The geodesic distance between any two nodes $v_a, v_b$  is computed as the accumulated edge weights along their shortest path on the graph $\mathcal{G}$ \cite{Wang2015,Wang2017saliency}:
\begin{equation}
    \begin{aligned}
    d_{geo}(v_a, v_b)  = \min_{\mathcal{C}_{v_a, v_b}}\sum\nolimits_{p,q} w_{pq}, ~~~~~p,q \in \mathcal{C}_{v_a, v_b},
    \end{aligned}
    \label{eq:3}
\end{equation}
where $\mathcal{C}_{v_a, v_b}$ is a path connecting the nodes $v_a, v_b$.

For frame $I^{t-1}$, the superpixels $\textbf{Y}^{t-1}$ can be decomposed into foreground regions $\textbf{F}$ and background regions $\textbf{B}$ according to the segmentation $M^{t-1}$, where $\textbf{Y}^{t-1} = \textbf{F} \cup \textbf{B}$.\footnote{The segmentation mask $M^{t-1}$ is pixel-level, we consider the superpixel contains more (less) foreground pixels than background ones as foreground (background).} Based on the graph $\mathcal{G}$, we use geodesic distance to define the similarity for superpixel $Y_i^{t}$ of frame $I^{t}$ to foreground regions $\textbf{F}$ and background regions $\textbf{B}$:
  \begin{equation}
    \begin{aligned}
    &D_F(Y_i^{t})~~~~= \min_{v_f\in\textbf{F}}d_{geo}(Y_i^{t}, v_f),\\
    &D_B(Y_i^{t})~~~~= \min_{v_b\in\textbf{B}}d_{geo}(Y_i^{t}, v_b).
    \end{aligned}
    \label{eq:4}
 \end{equation}
If a superpixel is close to the foreground (background), there exists a relatively short path to the foreground (background) nodes, and the value of $D_F$ ($D_B$) is small. For superpixel $Y_i^{t}$ of frame $I^{t}$, its dynamic foreground probability using geodesic distance is computed as:
\begin{equation}
    \begin{aligned}
    \textsl{G}_{dynamic}(Y_i^{t})= \frac{D_B(Y_i^{t})}{D_F(Y_i^{t})+D_B(Y_i^{t})}.
    \end{aligned}
    \label{eq:5}
\end{equation}

Our geodesic distance based confidence map is facilitated by the segmentation of prior frame and evaluated in a frame-by-frame fashion. We find the dynamic global model captures variations across the frames accurately, separates object from the background clearly, and complements pyramid model nicely. Therefore, we combine these two models. The final foreground probability of a superpixel $Y_i^{t}$ is:
\begin{equation}
    \begin{aligned}
    \textsl{G}(Y_i^t)= \textsl{G}_{static}(Y_i^t)\cdot \textsl{G}_{dynamic}(Y_i^t).
    \end{aligned}
    \label{eq:6}
\end{equation}
An example for the integration of dynamic and static confidence maps is presented in Fig.~\ref{fig:globalvisual}.

\subsection{Local Classifier by Propagation Uncertainty}
\label{sec:localclassifier}
\begin{figure*}[t]
  \center
      \includegraphics[width=1.0 \linewidth]{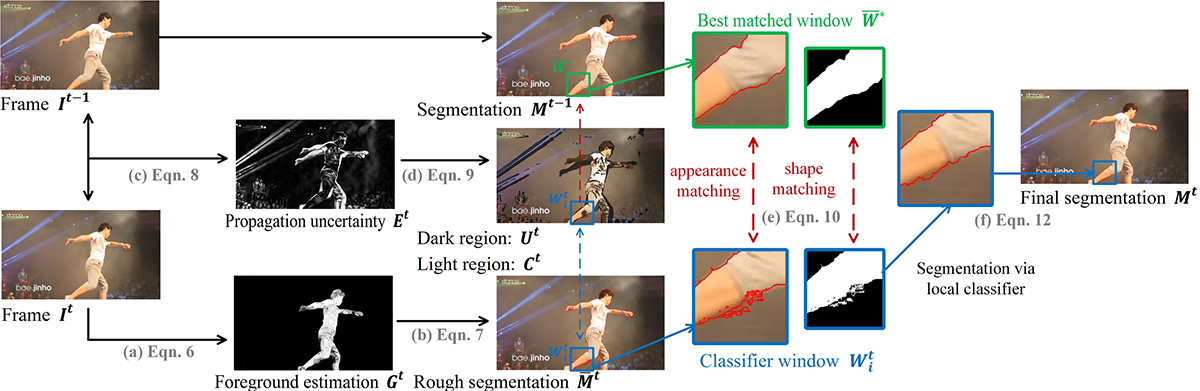}
\caption{Overview of the segmentation process. See ordered steps (a) through (f).}
  \label{fig:Overview}
\end{figure*}
After computing global confidence maps, we obtain a coarse segmentation mask $\M$ (see Fig.~\ref{fig:globalvisual} (e)):
\begin{equation}
    \begin{aligned}
    \M(\textbf{x}) = \left\{
        \begin{aligned}
            &~1~~~~~~~~~~\text{if } ~~\textsl{G}^t(\textbf{x}) > \text{mean}(\textsl{G}^t); \\
            &~0~~~~~~~~~~\text{else}.
        \end{aligned}
    \right.
    \end{aligned}
    \label{eq:7}
\end{equation}
for each pixel $\textbf{x}\!=\!(x,y)$. Obviously, this thresholding strategy is not always reliable. To resolve segmentation ambiguities, we adopt local classifiers on local image features. Previous methods, \textit{e.g.}, \cite{Bai2009}, track classifier windows along the foreground boundaries, which is computationally expensive due to motion estimation and sensitive to topological changes. Instead, we determine ambiguous regions and apply local classifiers. To this end, we use propagation uncertainty estimation:
 \begin{equation}
    \begin{aligned}
    E^t(\textbf{x}) = \| I^{t-1}(\textbf{x}) - I^{t}(\textbf{x})\|,
    \end{aligned}
    \label{eq:8}
\end{equation}
as the difference between two successive frames. This measurement is simple yet effective. The intuition is straightforward, the variations of appearance usually accompany the changes of label and potentially lead ambiguity. We separate the pixels in frame $I^t$ into two parts: \textit{propagation-uncertainty set} $\textbf{U}^t$ and \textit{propagation-certainty set} $\textbf{C}^t$:
\begin{equation}
\begin{aligned}
    \textbf{U}^t \!\!=\! \{\textbf{x}~\!|~\!\M(\textbf{x})\! \neq \!M^{t-1}(\textbf{x})\} \cup \{\textbf{x}~\!|~\!E^t(\textbf{x})\! > \!\text{mean}(E^t) \},\\
    \textbf{C}^t \!\!=\! \{\textbf{x}~\!|~\!\M(\textbf{x})\! = \!M^{t-1}(\textbf{x})\} \cap \{\textbf{x}~\!|~\!E^t(\textbf{x})\! \leq \!\text{mean}(E^t) \}.
    \end{aligned}
    \label{eq:9}
\end{equation}

The propagation-uncertainty set $\textbf{U}^t$ consists of all the pixels with changed labels and the pixels with high propagation uncertainty. The labeling in the propagation-certainty $\textbf{C}^t$ are relatively reliable, as their labels are consistent and appearance differences are small for two successive frames.

Then we sample a set of overlapping classifier windows $\{W^t_1,W^t_{2},{\cdots}\}$ cover all the image frame domain, where the neighboring windows overlap for half-window size. For frame $I^t$, we only enable the local classifiers whose classification window covers the propagation-uncertainty area $\textbf{U}^t$. We denote the window lattice as $\Omega$. For each window $W^t$ with its segments $\M_W$, we find its \textit{best matched} window $\overline{W}^*$ in frame $I^{t-1}$ via a match score using shape and color information:
\begin{equation}
   \begin{aligned}
        \!\!\!\!\overline{W}^*\!\!\! = \! \mathop{\argmin}_{\overline{W}} \sum_{\textbf{x}\in\Omega}\underbrace{|\M_W(\textbf{x}) \!-\! M^{t\!-\!1}_{\overline{W}}(\textbf{x})|}_{shape~matching} + \!\!\!\!\underbrace{|W^t(\textbf{x}) \!-\! \overline{W}(\textbf{x})|}_{appearance~matching},
     \end{aligned}
    \label{eq:10}
\end{equation}
where $\overline{W}(\textbf{x})$ indicates the pixel color in position $\textbf{x}$ of the window $\overline{W}$. We restrict the matching process inside a $h \times w$ search area, where $h$ ($w$) is set to the one fourth of the image frame's height (width). Using this function, the classification window $W^t_i$ is aligned with a best-matched window of the previous frame. We establish a discriminative classifier $L^t$:
\begin{equation}
    \begin{aligned}
    L^t(\textbf{x}) =  M^{t-1}(\textbf{x}^*), \textbf{x}\in W^t, \textbf{x}^* \in \overline{W}^*,
    \end{aligned}
    \label{eq:11}
\end{equation}
where $\textbf{x}^*$ corresponds to the nearest pixel of $\textbf{x}$ in $\overline{W}^*$, measured via color similarity. A pixel within propagation-uncertainty set $\textbf{U}^t$ is assigned to the label of the nearest pixel of its best matched window.

Overall, our local classifier integrates patch-level features such as shape and appearance information, and pixel-level matching. In our experiments, we observe such best-matching strategy based classifier is more effective than traditional color statistics based appearance models, especially for ambiguous color distributions.

Finally, we derive our pixel-wise foreground segmentation mask $M^t$ for frame $I^t$ from the coarse
segmentation mask $\M$ and local classifiers $L^t$:
\begin{equation}
    \begin{aligned}
    M^t(\textbf{x}) = \left\{
        \begin{aligned}
            &\M(\textbf{x})~~~~~~~~~~~~~~~~\!\text{if }~\textbf{x}\! \in \!\textbf{C}^t; \\
            &L^t(\textbf{x})~~~~~~~~~~~~~~~~~\text{if }~\textbf{x}\! \in \!\textbf{U}^t.
        \end{aligned}
    \right.
    \end{aligned}
    \label{eq:12}
\end{equation}
Then we propagate mask $M^t$ frame-by-frame in forward direction via our global and local classifiers described above.

So far, we described a forward propagation workflow, where we forward propagate the initial segmentation from an initial starting frame (see Fig. \ref{fig:Overview}).
Recall our approach accepts $K$ annotation frames $\widehat{\textbf{I}} = \{I^{n_1},I^{n_2},{\cdots}, I^{n_K}\}$ as input, the segmentation is proceeded in a bi-directional workflow. Let $(l_t, r_t)$ be the indices of the closest labeled frames before and after frame $I^t$, respectively (`left' and `right' of $I^t$).\footnote{The indices $(l_t, r_t)$ might not exist if $t<n_1$ or $t>n_K$. For clarity we omit such cases, since they do not affect the method description.} Our method processes cutout for frame $I^t$ starting from its closest labeled frames on either side in both the forward and backward directions. For a frame, there are two foreground masks
$M^t_{left}$ and $M^t_{right}$ computed in forward and backward processes, respectively. We merge these estimations from two processes via:
\begin{equation}
    \begin{aligned}
    M^t = \sigma\big(\frac{(r_t-t)\cdot M^t_{left} + (t-l_t)\cdot M^t_{right}}{r_t-l_t}\big),
    \end{aligned}
    \label{eq:13}
\end{equation}
where $\sigma()$ is a boolean function that returns 0 if the input less than 0.5, otherwise returns 1. Finally, morphological operations are adopted for filling the small holes of the final segmentation $M^t$.

\subsection{Intelligent Annotation Frame Selection}
\label{sec:intelligentannotationframeselection}
\begin{figure*}
  \centering
      \includegraphics[width=0.9 \linewidth]{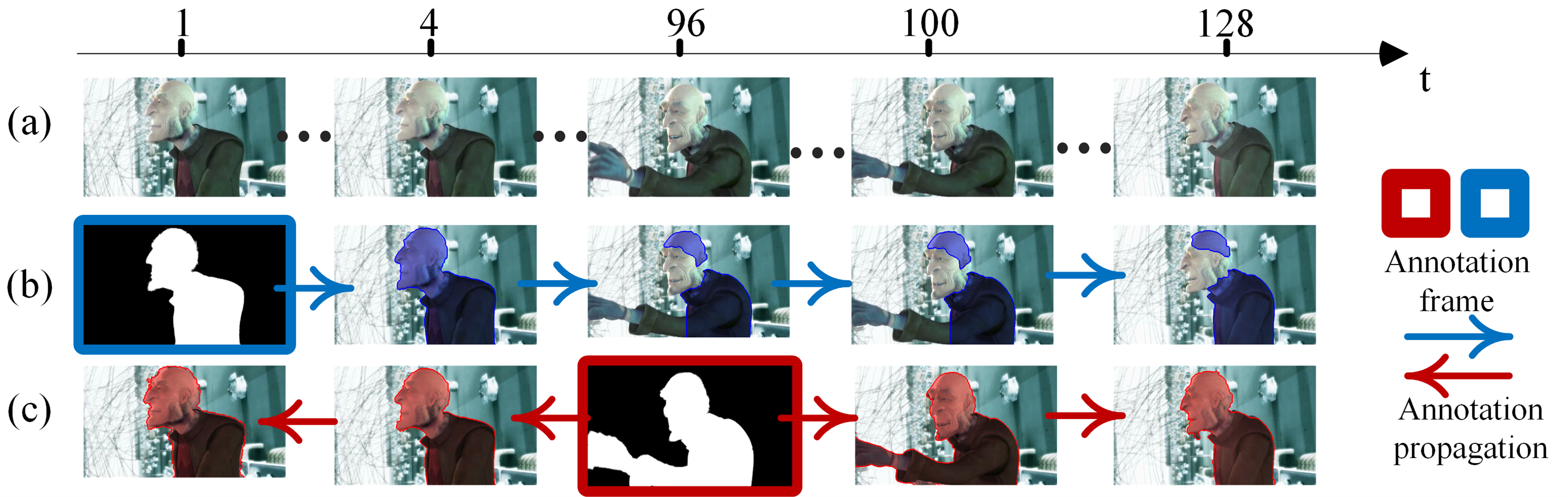}
\caption{Automatically selecting the most `informative' frames for annotation. (a) Input video sequence. (b) Traditional cutout method like \cite{Bai2009} transfers the mask of arbitrarily selected frames, which easily loose important information among the whole video sequence, such as the arm of the human. (c) Our approach selects more useful frames for human labeling, thus obtaining better results and alleviating the burden on the user.}
  \label{fig:propgation}
\end{figure*}

Previous cutout systems arbitrarily select frames with initial annotations (the first frame or key frames), which cannot ensure optimal propagation of user annotations. Besides, from the interaction point of view, it is preferable if the user is asked to annotate the frames that would best benefit the segmentation performance (instead of being forced to label random or less informative frames). To this end, we introduce an intelligent strategy to select the frames for user annotation where propagation uncertainty guides the selection of the frames for manual labeling. For offering a quick answerback, our annotation frame recommendation system is based on a simplified model of our propagation method, adopting superpixel as basic elements. We simplify the local uncertainty propagation as a region matching problem of finding the closest region in the neighboring frame. The match score is computed as the color difference between two superpixels.

Inspired by \cite{Sudheendra2012}, we model the probability that a region is mislabeled due to fail matching. We define a propagation error for a prior frame $I^{t-k}$ forward to frame $I^{t}$ via a probability model. All terms are analogously defined for propagating from a later frame $I^{t+k}$. The probability of superpixel $Y^t_i$ will be mislabeled when we obtain its label from frame $I^{t-1}$ is:
\begin{equation}
    \begin{aligned}
    \!\!\!\!\!Q(Y^t_i\!,t\!-\!1,t)\!=\!1\!-\!\exp\!\big(\!\!-\!(d_{app}(Y^t_i\!,t)\!+\!d_{occ}(Y^t_i\!,t) )\big).
    \end{aligned}
    \label{eq:14}
\end{equation}
The component distances reflect the expected propagating error. The term $d_{app}$ computes the color difference between region $Y^t_i$ and the best matched region $Y^{t-1}_{i'}$ in frame $I^{t-1}$:
\begin{equation}
    \begin{aligned}
    d_{app}(Y^t_i,t) = \|\mu(Y^t_i) - \mu(Y^{t-1}_{i'})\|,
    \end{aligned}
    \label{eq:15}
\end{equation}
where $\mu(\cdot)$ represents the mean color of the superpixel.

The term $d_{occ}$  measures occlusions using the consistency of the forward and backward matching:
\begin{equation}
    \begin{aligned}
    d_{occ}(Y^t_i,t) = \frac{ \|\overrightarrow{\textbf{f}}(Y^{t-1}_i) + \overleftarrow{\textbf{f}}(Y^{t}_{i'})\|}{\|\overrightarrow{\textbf{f}}(Y^{t-1}_i)\|+\|\overleftarrow{\textbf{f}}(Y^{t}_{i'})\|},
    \end{aligned}
    \label{eq:16}
\end{equation}
where $\overrightarrow{\textbf{f}}(\cdot)$ is the vector from the center of a superpixel to the center of its best matched superpixel in next frame, similarly, $\overleftarrow{\textbf{f}}(\cdot)$ indicates the vector from the center of a superpixel to the center of its best matched superpixel in prior frame.
Ideally, if two superpixels $Y^t_i$ and $Y^{t-1}_{i'}$ are matched, we expect an one-to-one correlation and
these two flows should be opposite in direction, making the numerator close to 0. 

When there is more than one frame between labeled frame $I^{l_t}$ and current frame $I^t$, we predict errors accumulated over successive frames. Defining the error recursively, we have:
\begin{equation}
\begin{aligned}
    Q(Y^t_i\!,t\!\!-\!\!j,t)&\!=\!Q(Y^t_i,t\!-\!j\!+\!1,t)  \\
    &\!+\!(1\!-\!Q(Y^t_i,t\!-\!j\!\!+\!\!1,t))Q(Y^{t\!-\!j\!+\!1}_{i'}\!,t\!-\!j,t\!-\!j\!+\!1),
    \end{aligned}
    \label{eq:17}
\end{equation}
where $Y^{t\!-\!j\!+\!1}_{i'}$ indicates the matched superpixel of $Y^t_i$ in frame $I^{t\!-\!j\!+\!1}$. In other words, superpixel of $Y^t_i$ was either mislabeled along some hope from $I^{t\!-\!j\!+\!1}$ forward to $I^t$, or else those hopes were all correct and the wrong label was propagated from the single hop from adjacent frames $I^{t\!-\!j}$ and $I^{t\!-\!j\!+\!1}$.

Then we derive a $N\times N$ propagation error matrix, where $(t',t)$-th element indicates the total propagation error from $t'$-th frame to $t$-th frame:
\begin{equation}
    \begin{aligned}
    \mathcal{E}(t',t) = \alpha(t',t)\cdot\sum\nolimits_{i}|Y^t_i| \cdot Q(Y^t_i,t',t),
     \end{aligned}
    \label{eq:18}
\end{equation}
where $|\cdot|$ is the number of the elements in the collection.

For modeling our global pyramid appearance estimation, we introduce an adjustment coefficient $\alpha(t',t)$, which represents the difficulty of propagating labels from frame $I^{t'}$ to frame $I^{t}$ according to their difference on frame level, which is computed as:
\begin{equation}
    \begin{aligned}
    \alpha(t',t) = \sum\nolimits_{\ell = 0}^{L} ||H_{t'}^\ell - H_{t}^\ell ||,
     \end{aligned}
    \label{eq:19}
\end{equation}
where $H_{t}^\ell$ is our pyramid appearance representation of frame $I^{t}$ at $\ell$ level. $\alpha(t',t)$ measures the appearance similarities of two frames with location information.

We find that, the more similarities of two frames, the higher propagation accuracy we can achieve. This observation is intuitive; when two frames are similar in their pyramid appearance representation, our pyramid model would performance well. Therefore, this term contributes to selecting the `discriminative' frames so that our global pyramid appearance model can offer more reliable prediction.

We define the optimization problem for selecting the best set of frames from which to
propagate, aiming to choose $\widehat{\textbf{I}}^{~*} = \{I^{n_1},I^{n_2},{\cdots}, I^{n_K}\}$ by minimizing the total expected number of erroneous pixels in entire video:
\begin{equation}
    \begin{aligned}
    \widehat{\textbf{I}}^{~*}\!\!\! =  \mathop{\argmin}_{\widehat{\textbf{I}}~\subset~\textbf{I}} \sum \nolimits_{t = 1}^{N} \frac{(r_t\!-\!t)\!\cdot\!\mathcal{E}(l_t,t) + (t\!-\!l_t)\!\cdot\! \mathcal{E}(r_t,t)}{r_t-l_t},
     \end{aligned}
    \label{eq:20}
\end{equation}

This propagation error model is derived from our propagation strategy in Sec.~\ref{sec:localclassifier}. By above minimization, we select $K$ number of useful frames according to the expected label error. Here, we set $K$ as a manually selected parameter.
For every frame $I^t$,  we compute its predicted mislabeling rate when propagating the labels from frame $I^{t'}$, according to the expected propagation error and dissimilarity between frames. We then formulate such annotation frame selection as an optimization problem to minimize total propagation error.
Eqn.~\ref{eq:20} can be efficiently solved by dynamic programming algorithm, similar to \cite{Sudheendra2012}.
In this way, our method reduces total manual effort by keeping the number of selected frames low.
We illustrate the process of our intelligent annotation frame selection in Fig. \ref{fig:propgation}. Traditional video segmentation methods arbitrarily select annotation frames, as Fig.\ref{fig:propgation} (b). While the selected frames maybe miss a part of important regions, which easily leads to unreliable segments in other frames. Our method tries to find the most `informative' frames for labelling, thus maximizing the human labor and obtaining better results (see the segments of 100-th and 128-th frames in Fig.\ref{fig:propgation} (c)). We summarize our whole video cutout method in Alg. \ref{alg:alg1}.

\begin{algorithm}
\caption{Selective Video Cutout}
\label{alg:alg1}
\begin{algorithmic}[1]
\Require Input video sequence $\textbf{I} = \{I^{1},I^{2},{\cdots}, I^{N}\}$, number of annotation frames $K$;
\Ensure Binary segmentation masks $\textbf{M} = \{M^{1},{\cdots}, M^{N}\}$;

\text{\!\!\!\!\!\!\!\!\!\!\!/*~Get User Annotations */}
\State Choose annotation frames $\widehat{\textbf{I}} = \{I^{n_1},{\cdots}, I^{n_K}\}$ via (\ref{eq:20});
\State Obtain binary masks $\widehat{\textbf{M}} = \{M^{n_1},M^{n_2},{\cdots}, M^{n_K}\}$ via user annotation;
\State Obtain static confidence ${G^t_{static}}_t$ via (\ref{eq:1});

\text{\!\!\!\!\!\!\!\!\!\!\!/*~Forward Propagation Workflow */}
\For  {$t = n_1:N$}
\If  {$I^{t} \notin\widehat{\textbf{I}}$}
\State Obtain dynamic confidence $G^t_{dynamic}$ via (\ref{eq:5})
$~~~~~~~~~~$from previous frame $I^{t-1}$;
\State Obtain global confidence $G^t$ via (\ref{eq:6});
\State Obtain coarse segmentation $\overline{M}^t$ via (\ref{eq:7});
\State Obtain segmentation mask $M_{left}^t$ via (\ref{eq:12});
\EndIf
\EndFor

\text{\!\!\!\!\!\!\!\!\!\!\!/*~Backward Propagation Workflow */}
\For  {$t = n_K:1$}
\If  {$I^{t} \notin \widehat{\textbf{I}}$}
\State Obtain dynamic confidence $G^t_{dynamic}$ via (\ref{eq:5})
$~~~~~~~~~~$from next frame $I^{t+1}$;
\State Obtain global confidence $G^t$ via (\ref{eq:6});
\State Obtain coarse segmentation $\overline{M}^t$ via (\ref{eq:7});
\State Obtain segmentation mask $M_{right}^t$ via (\ref{eq:12});
\EndIf
\EndFor

\text{\!\!\!\!\!\!\!\!\!\!\!/*~Merge Results from Two Processes */}
\State Obtain final segmentation masks $\textbf{M}$ via (\ref{eq:13});
\end{algorithmic}
\end{algorithm}
\begin{table*}[t]
\centering
\caption{Region similarity ($\mathcal{J}$) and contour accuracy ($\mathcal{F}$) on a representative subset of
the DAVIS dataset \cite{Perazzi2016}, and the average computed over all 50
video sequences. For both two measure metrics, higher values are better. The best results are boldfaced, the second best is in blue.}
\setlength\tabcolsep{4pt}
\renewcommand\arraystretch{1.1}
\resizebox{1\textwidth}{!}{

\begin{tabular}{|c|c||c|c|c|c|c|c|c||c|c|c|c|c|c|c|}  
\hline
\multirow{2}*{Dataset}
&\multirow{2}*{Video}
&\multicolumn{7}{c||}{Region Similarity $\mathcal{J}$} & \multicolumn{7}{c|}{Contour Accuracy $\mathcal{F}$}  \\
\cline{3-16}
&&BV16 &FP15 &JC15 &SS14 &TS13 &HV10 &Ours &BV16 &FP15 &JC15 &SS14 &TS13 &HV10 &Ours\\
\hline
\hline
\multirow{21}*{\textit{DAVIS}}
&blackswan	    &0.943   &0.908    &0.930    &0.932    &0.871    &0.916    &\textbf{0.945}  &0.965    &0.904    &0.945    &0.956    &0.856    &0.910    &\textbf{0.979}\\
&breakdance	    &0.499   &0.566    &0.477    &0.329    &0.056    &0.549    &\textbf{0.614} &0.487    &0.472    &0.511    &0.389    &0.069    &0.473    &\textbf{0.605}\\
&car-roundabout &0.850   &0.716    &0.725    &0.707    &0.613    &0.777   &\textbf{0.864}  &0.624    &0.478    &0.618    &\textbf{0.709}   &0.434    &0.551    &0.706\\
&car-shadow	    &0.578   &0.722    &0.645    &\textbf{0.774}    &0.635    &0.699    &0.744  &0.473    &0.642    &0.624    &\textbf{0.755}    &0.512    &0.594    &0.715\\
&car-turn	    &0.844   &0.723    &0.834    &\textbf{0.908}     &0.322    &0.809   &0.871  &0.689    &0.614    &0.742    &\textbf{0.882}    &0.378    &0.604    &0.786\\
&drift-chicane	&0.033   &\textbf{0.456}    &0.243    &0.119    &0.018    &0.330    &0.427  &0.076    &0.476    &0.338    &0.159    &0.033    &\textbf{0.546}    &0.540\\
&goat	        &0.661   &0.677    &\textbf{0.730}    &0.535    &0.444    &0.579    &0.709  &0.583    &0.576    &0.617    &0.469    &0.403    &0.545    &\textbf{0.631}\\
&hike	        &0.755   &0.874    &0.663    &0.775    &0.679    &0.876    &\textbf{0.881}  &0.764    &0.911    &0.743    &0.795    &0.675    &0.877    &\textbf{0.925}\\
&horsejump-low	&0.601   &0.606    &\textbf{0.662}    &0.497    &0.291    &0.551    &0.652  &0.564    &0.533    &\textbf{0.696}    &0.547    &0.355    &0.572    &0.673\\
&kite-walk	    &\textbf{0.869}   &0.681    &0.509    &0.498    &0.446    &0.765    &0.845  &\textbf{0.728}    &0.411    &0.359    &0.354    &0.435    &0.623    &0.662\\
&mallard-fly	&\textbf{0.606}   &0.541    &0.536    &0.556    &0.200    &0.436    &0.561  &\textbf{0.645}    &0.538    &0.579    &0.607    &0.234    &0.441    &0.593\\
&motocross-bumps&0.400   &0.306    &\textbf{0.760}    &0.470    &0.133    &0.534    &0.653  &0.490    &0.302    &\textbf{0.743}    &0.519    &0.183    &0.548    &0.647\\
&paragliding	&0.875   &0.866    &\textbf{0.951}    &0.863    &0.734    &0.907    &0.927  &0.773    &0.727    &\textbf{0.906}    &0.744    &0.634    &0.856   &0.904\\
&rollerblade	&0.587   &0.449   &0.725    &0.138    &0.097    &0.460    &\textbf{0.734}	 &0.644    &0.576    &\textbf{0.759}    &0.155    &0.142    &0.552    &0.738\\
&scooter-black	&0.337   &0.503    &0.626    &\textbf{0.792}    &0.377    &0.624    &0.778  &0.407    &0.363    &0.528    &\textbf{0.721}    &0.410    &0.574    &0.640\\
&soccerball	    &0.844   &0.819    &0.096    &0.652    &0.028    &0.065   &\textbf{0.866}  &0.849    &0.835    &0.127    &0.696    &0.059    &0.074   &\textbf{0.917}\\
&stroller	    &0.766   &0.597    &0.656    &0.464    &0.369    &0.662    &\textbf{0.842}  &0.790    &0.581    &0.718    &0.525    &0.404    &0.707    &\textbf{0.866}\\
&surf	        &0.491   &0.842    &\textbf{0.941}    &0.820    &0.814    &0.759    &0.912  &0.530    &0.712    &\textbf{0.871}    &0.732    &0.641    &0.652    &0.849\\
&swing	        &\textbf{0.784}   &0.648    &0.115    &0.510    &0.097    &0.104    &0.783  &\textbf{0.746}    &0.538    &0.109    &0.408    &0.087    &0.091    &0.726\\
&tennis 	    &0.737   &0.623    &0.764    &0.481    &0.074    &0.575    &\textbf{0.815}  &0.844    &0.652    &0.818    &0.536    &0.113    &0.578    &\textbf{0.872}\\

\cline{2-16}
&Avg.          &0.665   &0.631    &0.607    &0.556    &0.358    &0.596    &\textbf{0.688}  &0.656    &0.546    &0.586    &0.533    &0.346    &0.576    &\textbf{0.665}\\
\hline
\end{tabular}
}
\label{table1}
\end{table*}

\section{Experimental Evaluations}
\label{sec:results}
In this section, we first evaluate the overall performance by comparing to several state-of-the-art methods (Sec.~\ref{sec:PerformanceComparison}). To gain a deeper insight of our method, we study important parameter and access the performance of the pyramid model by comparisons to several baselines (Sec.~\ref{sec:PPC}). We further validate the effectiveness of annotation frame selection strategy via conducting user study (Sec.~\ref{sec:US}). Finally, runtime analysis is presented in Sec.~\ref{sec:RA}.

Our evaluations are on two benchmarks: the JumpCut dataset \cite{Fan2015} and recently released DAVIS dataset \cite{Perazzi2016}. JumpCut contains five sets of video clips (\textit{SNAPCUT}, \textit{ANIMAL}, \textit{HUMAN}, \textit{STATIC}, and \textit{FAST} sets) provided by \cite{Fan2015, Bai2009, Zhong2012}. The \textit{SNAPCUT} set contains three example videos from \cite{Bai2009} and the \textit{ANIMAL}, \textit{HUMAN}, and \textit{STATIC}, sets are collected from \cite{Zhong2012}. There are 22 video clips in total and full pixel-level segmentation ground-truth for each frame is available. Test data spans a wide degree of difficulty, such as highly complex color distributions, cluttered and dynamic backgrounds, and fast motion patterns. The DAVIS dataset consists of 50 videos, accompanied by per-frame and pixel-level ground-truth masks. Those videos  cover various common challenges in video segmentation, such as occlusions, dramatic topology changes, and large appearance variations.

\begin{figure*}
  \centering
      \includegraphics[width=1.0 \linewidth]{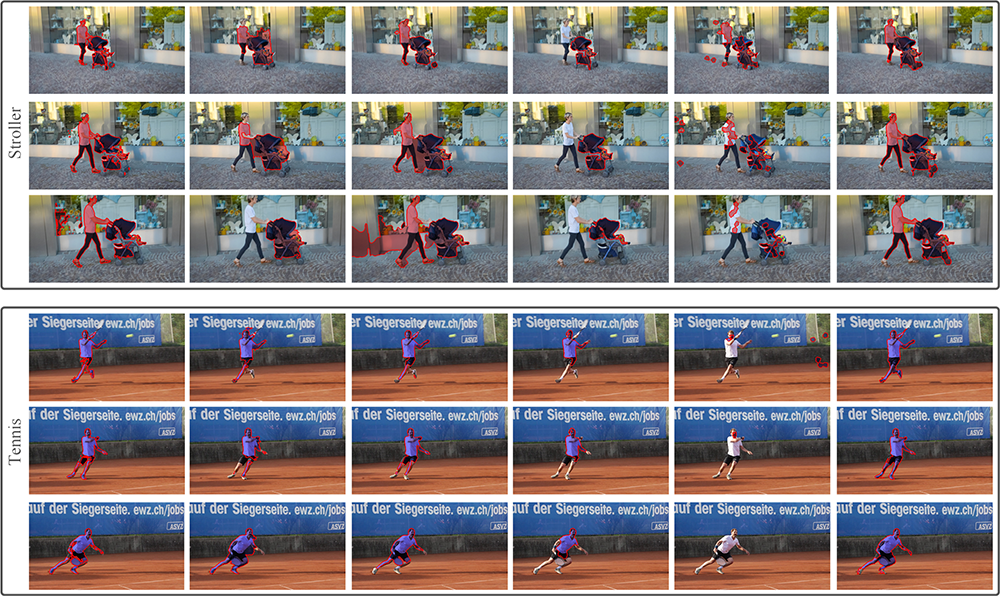}
        \\ \mbox{}\hfill\mbox{} ~~~~(a) \hfill\mbox{}
     \mbox{}\hfill (b)  \hfill\mbox{}
     \mbox{}\hfill (c) \hfill\mbox{}
     \mbox{}\hfill (d) \hfill\mbox{}
     \mbox{}\hfill \!(e) \hfill\mbox{}
     \mbox{}\hfill (f) \hfill\mbox{}
\caption{Qualitative comparison against the state-of-the-art methods on two video sequences from the DAVIS benchmark \cite{Perazzi2016} (\textit{stroller} and \textit{tennis}). From left to right: (a) BV16 \cite{maerki2016}, (b) FP15 \cite{Perazzi2015}, (c) JC15 \cite{Fan2015}, (d) SS14 \cite{Ramakanth2014}, (e) TS13 \cite{chang2013}, (f) Our method.}
  \label{fig:exp_ComparisonDAVIS}
  \mbox{}\hfill
\end{figure*}

\subsection{Performance Comparison}
\label{sec:PerformanceComparison}
We first evaluate our approach with respect to existing methods on the task of foreground mask propagation, which is the core of our video cutout technique. These experiments are conducted by comparing with the state-of-the-art alternatives, including the Rotobrush of Adobe AfterEffects (RB09), which is based on the Video SnapCut~\cite{Bai2009}, discontinuity-aware video cutout (DA12)~\cite{Zhong2012}, and JumpCut (JC15)~\cite{Fan2015} using non-successive mask transfer. These three methods are designed for video cutout purposes. Additionally, we compare to five very recent segmentation tracking methods: BV16~\cite{maerki2016}, FP15~\cite{Perazzi2015}, SS14~\cite{Ramakanth2014}, TS13~\cite{chang2013}, and HV10~\cite{grundmann2010}. For each test video sequence, we compare the segmentation performance with manual annotation of the first frame as initialization. To keep our analysis fair, all the methods predict subsequent frames without any additional user input and our method only adopts forward propagation workflow.

\subsubsection{Performance on DAVIS Dataset}
We evaluate the effectiveness of our approach on DAVIS dataset \cite{Perazzi2016} with two accompanied evaluation tools: intersection-over-union metric ($\mathcal{J}$) for measuring the region-based segmentation similarity and F-measure ($\mathcal{F}$) for measuring the contour accuracy.  Given a segmentation mask $M$ and ground-truth $G$, IoU score is defined as
\begin{equation}
    \begin{aligned}
    \mathcal{J} = \frac{M\bigcap G}{M\bigcup G}.
     \end{aligned}
\end{equation}
Contour accuracy ($\mathcal{F}$) is for measuring how well the segment contours $c(M)$ match the ground-truth contour $c(G)$. Contour-based precision $P_c$ and recall $R_c$ between $c(M)$ and $c(G)$ can be computed via bipartite graph matching. Given $P_c$ and $R_c$ , contour accuracy $\mathcal{F}$ is computed as
\begin{equation}
    \begin{aligned}
     \mathcal{F} = \frac{2P_cR_c}{P_c+R_c}.
     \end{aligned}
\end{equation}

The results for representative sequences and the average performance over the \textit{entire} DAVIS dataset are reported in Table~\ref{table1}.
As can be seen, our approach performs better than all other methods overall,
achieving the best IoU score $\mathcal{J}$ on most of the videos with the average score up to $\textbf{0.688}$. Similar conclusions can be drawn from the contour accuracy $\mathcal{F}$ where our approach also achieves the highest overall score ($\textbf{0.665}$).
Qualitative comparison results for DAVIS dataset are shown in Fig.~\ref{fig:exp_ComparisonDAVIS}, which demonstrate the superiority of the proposed method on challenging scenarios.

\begin{table*}
\caption{Error rates for automatic mask propagation across different frame distances on the JumpCut dataset \cite{Fan2015} for
 several different methods: RB09~\cite{Bai2009}, DA12~\cite{Zhong2012}, SS14~\cite{Ramakanth2014} and JC15~\cite{Fan2015}. Lower values are better. The best results are boldfaced.}
\setlength\tabcolsep{2pt}
\renewcommand\arraystretch{1.1}
\resizebox{1\textwidth}{!}{
\begin{tabular}{|c|c||c|c|c|c|c||c|c|c|c|c||c|c|c|c|c||c|c|c|c|c|}  
\hline
\multirow{2}*{Dataset}
&\multirow{2}*{Video}
&\multicolumn{5}{c||}{1} & \multicolumn{5}{c||}{4} & \multicolumn{5}{c||}{8} & \multicolumn{5}{c|}{16}\\
\cline{3-22}
&&RB09 &DA12 &SS14 &JC15 &Ours &RB09 &DA12 &SS14 &JC15 &Ours &RB09 &DA12 &SS14 &JC15 &Ours &RB09 &DA12 &SS14  &JC15 &Ours\\
\hline
\hline
\multirow{5}*{\textit{ANIMAL}}
&bear      &1.82 &1.07 &1.84 &1.36 &\textbf{0.86} &6.74 &1.83 &2.93 &2.91 &\textbf{1.47} &5.58 &2.72 &3.44 &3.18 &\textbf{1.93} &4.58 &4.48 &4.21 &4.00 &\textbf{2.11}\\
&giraffe  &8.49 &6.99 &4.77 &3.83 &\textbf{3.57} &12.7 &8.08 &8.82 &6.11 &\textbf{5.14} &14.4 &9.27 &11.6 &\textbf{6.69} &7.80 &22.0 &11.2 &17.4 &\textbf{7.40} &9.67\\
&goat  &3.68 &2.57 &3.30 &2.00 &\textbf{1.91} &7.00 &4.74 &5.45 &3.35 &\textbf{3.32} &10.6 &8.61 &6.34 &\textbf{3.41} &4.13 &13.1 &13.3 &8.22 &\textbf{4.14} &4.97\\
&pig         &3.86 &\textbf{2.08} &3.39 &2.97 &2.37 &7.02 &3.74 &5.84 &3.27 &\textbf{2.80} &11.3 &7.08 &9.07 &3.58 &\textbf{2.92} &9.22 &9.85 &10.3 &3.43 &\textbf{3.24}\\
\cline{2-22}
&Avg.        &4.46 &3.18 &3.33 &2.54 &\textbf{2.17} &8.36 &4.59 &5.76 &3.91 &\textbf{3.18} &10.4 &6.92 &7.61 &4.22 &\textbf{4.19} &12.2 &9.70 &10.0 &\textbf{4.74} &4.99\\
\hline
\hline
\multirow{4}*{\textit{HUMAN}}
&couple      &4.09 &3.54 &3.78 &2.27 &\textbf{2.06} &10.0 &5.90 &12.7 &4.35 &\textbf{4.13} &18.1 &11.0 &17.2 &\textbf{4.81} &5.50 &17.5 &16.0 &23.4 &\textbf{5.13} &8.49\\
&park        &3.95 &3.49 &3.33 &2.93 &\textbf{1.95} &6.28 &4.14 &4.47 &5.06 &\textbf{3.46} &8.97 &4.60 &5.07 &5.19 &\textbf{4.40} &11.8 &6.54 &6.91 &5.39 &\textbf{5.33}\\
&station     &2.53 &2.01 &2.37 &1.53 &\textbf{1.48} &4.24 &6.66 &8.58 &3.82 &\textbf{3.52} &6.66 &14.0 &16.0 &6.80 &\textbf{6.13} &8.85 &20.9 &21.3 &9.01 &\textbf{8.42}\\
\cline{2-22}
&Avg.        &3.52 &3.01 &3.16 &2.24 &\textbf{1.83} &6.84 &5.56 &8.58 &4.41 &\textbf{3.71} &11.2 &9.86 &12.7 &5.60 &\textbf{5.35} &12.7 &14.4 &17.2 &\textbf{6.51} &7.41\\
\hline
\hline
\multirow{5}*{\textit{STATIC}}
&car         &1.35 &1.38 &0.73 &0.54 &\textbf{0.50} &1.43 &1.42 &1.45 &\textbf{0.87} &1.27 &1.52 &1.79 &2.18 &\textbf{0.99} &1.72 &\textbf{1.76} &5.93 &5.08 &2.26 &2.57\\
&cup         &3.72 &1.34 &1.19 &1.16 &\textbf{0.71} &4.14 &1.87 &1.61 &1.73 &\textbf{1.17} &5.16 &4.65 &3.02 &1.93 &\textbf{1.65} &5.45 &12.9 &9.31 &\textbf{2.15} &2.40\\
&pot         &0.94 &1.49 &0.80 &0.70 &\textbf{0.39} &1.58 &1.56 &1.05 &1.28 &\textbf{0.66} &1.71 &2.11 &1.41 &1.49 &\textbf{0.98} &2.43 &5.03 &2.98 &2.95 &\textbf{1.79}\\
&toy         &1.02 &1.32 &0.70 &0.58 &\textbf{0.31} &1.15 &1.45 &1.29 &1.18 &\textbf{0.69} &1.25 &2.05 &1.44 &1.23 &\textbf{0.97} &\textbf{1.28} &3.19 &2.16 &1.30 &1.49\\
\cline{2-22}
&Avg.        &1.75 &1.38 &0.85 &0.74 &\textbf{0.48} &2.07 &1.57 &1.35 &1.26 &\textbf{0.95} &2.41 &2.65 &2.01 &1.41 &\textbf{1.23} &2.73 &6.76 &4.88 &2.16 &\textbf{2.06}\\
\hline
\hline
\multirow{4}*{\textit{SNAPCUT}}
&animation &1.98 &1.26 &1.83 &1.59 &\textbf{0.80} &5.18 &3.43 &3.52 &4.46 &\textbf{2.13} &12.9 &5.59 &7.45 &4.46 &\textbf{3.31} &11.9 &6.38 &6.78 &4.55 &\textbf{3.35}\\
&fish       &2.80 &1.97 &2.54 &1.80 &\textbf{1.40} &7.68 &5.87 &7.32 &5.66 &\textbf{2.80} &13.8 &9.25 &10.8 &7.61 &\textbf{4.75} &51.8 &21.7 &25.7 &17.5 &\textbf{7.67}\\
&horse      &3.99 &4.14 &3.00 &2.62 &\textbf{1.39} &5.18 &11.3 &12.6 &3.93 &\textbf{2.81} &7.43 &28.4 &25.9 &5.48 &\textbf{3.47} &8.39 &45.1 &37.8 &6.80 &\textbf{4.84}\\
\cline{2-22}
&Avg.      &2.92 &2.45 &2.45 &2.00 &\textbf{1.20} &6.01 &6.86 &7.81 &4.68 &\textbf{2.58} &11.3 &14.4 &14.7 &5.85 &\textbf{3.84} &24.0 &24.3 &23.4 &9.61 &\textbf{5.28}\\
\hline
\hline
\multirow{9}*{\textit{FAST}}
&bball       &1.55 &1.71 &1.90 &1.61 &\textbf{1.12} &6.86 &3.24 &2.97 &2.16 &\textbf{2.13} &10.0 &4.70 &4.95 &\textbf{2.75} &3.23 &18.4 &8.47 &8.89 &\textbf{3.90} &4.16\\
&cheetah     &7.17 &3.99 &5.07 &4.41 &\textbf{2.76} &13.4 &6.82 &6.06 &4.87 &\textbf{4.44} &21.1 &10.6 &7.47 &5.97 &\textbf{5.87} &31.5 &16.6 &7.68 &8.16 &\textbf{7.10}\\
&dance       &6.65 &9.19 &7.55 &6.62 &\textbf{5.57} &29.8 &17.7 &30.9 &12.4 &\textbf{12.2} &40.2 &37.9 &53.4 &\textbf{18.3} &23.1 &56.1 &50.8 &43.0 &\textbf{18.7} &26.5\\
&hiphop      &8.02 &4.62 &6.94 &3.37 &\textbf{3.26} &36.4 &19.3 &27.8 &\textbf{8.48} &11.0 &73.4 &32.0 &39.1 &\textbf{11.0} &19.0 &67.5 &51.1 &33.7 &\textbf{14.2} &21.9\\
&kongfu      &5.42 &3.71 &5.78 &3.28 &\textbf{1.49} &26.6 &18.4 &12.3 &5.30 &\textbf{2.34} &25.6 &24.8 &18.8 &6.59 &\textbf{2.85} &40.2 &40.8 &17.9 &8.00 &\textbf{3.77}\\
&skater      &6.33 &5.33 &5.09 &4.89 &\textbf{3.97} &11.5 &8.93 &8.84 &6.78 &\textbf{6.36} &25.4 &21.2 &11.7 &8.02 &\textbf{7.52} &38.7 &40.8 &29.6 &22.8 &\textbf{21.4}\\
&supertramp &14.7 &8.99 &17.4 &\textbf{6.17} &8.84 &52.7 &32.2 &35.4 &22.6 &\textbf{14.0} &76.8 &42.2 &41.7 &31.3 &\textbf{22.4} &129  &60.5 &57.4 &42.9 &\textbf{27.1}\\
&tricking    &42.2 &9.71 &11.9 &5.02 &\textbf{3.53} &31.3 &21.4 &29.0 &8.31 &\textbf{7.22} &48.4 &41.6 &46.6 &17.9 &\textbf{15.9} &79.4 &70.9 &35.8 &21.3 &\textbf{21.2}\\
\cline{2-22}
&Avg.        &11.5 &5.90 &7.70 &4.42 &\textbf{3.81} &26.0 &15.9 &19.1 &8.86 &\textbf{7.46} &40.1 &26.8 &27.9 &12.7 &\textbf{12.4} &57.6 &42.4 &29.2 &17.4 &\textbf{16.5}\\
\hline
\end{tabular}
}
\label{table2}
\end{table*}

\subsubsection{Performance on JumpCut Dataset}

Following the experimental setup in \cite{Fan2015}, we report the error rates of the different methods for automatically propagating segmentation mask across different frame distances in the five video sets from the JumpCut benchmark. The propagation errors are computed as the ratio between the wrongly classified areas of the transferred mask and the foreground area of the ground-truth mask.
For each video sequence, the first 128 frames are tested for automatically propagating a ground-truth mask from frame $t$ to $t+d$, for $t = 0, 16, {\cdots}, 96$, with different transfer distances $d \in \{1, 4, 8, 16\}$.

From the results in Table~\ref{table2}, it can be observed that our method outperforms all other methods, yielding the smallest error in the majority of the cases. We additionally observe that many methods degrade in quality over long sequences, as errors accumulate over time. In contrast, our method achieves better scores on long videos, experiencing less drift of the object region than the others.

\subsection{Validation of Our Approach}
\label{sec:PPC}
In this section, we offer detailed exploration for the
proposed approach in several aspects with the test videos of JumpCut dataset~\cite{Fan2015}. We test the values of important parameters, and verify
basic assumptions of the proposed algorithm.

\subsubsection{Parameter Validation}
\begin{table}[!htbp]
\caption{Parameter selection for size of local window with IoU score.}
\centering
\setlength\tabcolsep{2pt}
\resizebox{0.49\textwidth}{!}{
\begin{tabular}{|c|c|c|c|c|}
\hline
 &$20\times20$ &$30\times30$ &$50\times50$ &$\{30\times30, 50\times50\}$ \\
 \hline
IoU Score &0.703 &0.717 &0.721  &0.745\\
 \hline
 &$70\times70$  &$80\times80$ &$100\times100$ &$\{30\times30, 50\times50, 80\times80\}$ \\
 \hline
IoU Score &0.72 &0.722 &0.707  &0.755\\
\hline
\end{tabular}
}
\label{table3}
\end{table}
We study the influence of the size of the window in our local classifier in Sec. \ref{sec:localclassifier}. The size of the
windows can vary according to the size of the object. In Table~\ref{table3}, we explore different
choices for the size of the local window, and report their corresponding performance with IoU score on the test set of JumpCut dataset~\cite{Fan2015}.
According to Table~\ref{table3}, we choose $\{30\times30, 50\times50, 80\times80\}$ pixels as the size of the local windows for capturing multi-scale information.

\begin{table*}[t]
\centering
\caption{IoU scores and selection times with different annotation frame numbers in our user study. Best results are in boldface.}
\resizebox{1\textwidth}{!}{
\begin{tabular}{|c|c||c|c|c||c|c|c||c|c|c||c|c|c||c|c|c|}  
\hline
\multirow{2}*{\!\!\!\!method\!\!\!\!}
&\multirow{2}*{criteria}
&\multicolumn{3}{c||}{1} & \multicolumn{3}{c||}{2} & \multicolumn{3}{c||}{3} & \multicolumn{3}{c||}{4} & \multicolumn{3}{c|}{5}\\
\cline{3-17}
&&\!\!\!\!random\!\!\!\! &\!\!user\!\! &\!\!\textbf{Ours}\!\! &\!\!\!\!random\!\!\!\! &\!\!user\!\! &\!\!\textbf{Ours}\!\! &\!\!\!\!random\!\!\!\! &\!\!user\!\! &\!\!\textbf{Ours}\!\! &\!\!\!\!random\!\!\!\!  &\!\!user\!\! &\!\!\textbf{Ours}\!\! &\!\!\!\!random\!\!\!\! &\!\!user\!\! &\!\!\textbf{Ours}\!\! \\
\hline
\hline
\multirow{2}*{\!\!\textit{RB09}\!\!}
&\!\!\!\!selection time(s)\!\!\!\!         &-     &54.5  &-    &-     &59.3  &-   &-     &63.8  &-  &-     &71.6 &-   &-     &73.5 &-\\
&IoU score                  &\!\!0.704\!\! &\!\!0.741\!\! &-    &\!\!0.765\!\! &\!\!0.750\!\! &-   &\!\!0.786\!\! &\!\!0.802\!\! &-  &\!\!0.819\!\! &\!\!0.851\!\!&-   &\!\!0.831\!\! &\!\!0.862\!\! &-\\
\hline
\multirow{2}*{\!\!\textbf{Ours}\!\!}
&\!\!\!\!selection time(s)\!\!\!\!          &-     &51.2  &7.5   &-     &57.5   &7.6    &-     &61.5 &7.6 &- &68.1 &7.6 &- &75.9 &7.6\\
&IoU score                  &\!\!0.742\!\! &\!\!0.772\!\! & \!\!\textbf{0.789}\!\! &\!\!0.796\!\! &\!\!0.822\!\!  & \!\!\textbf{0.858}\!\!  &\!\!0.817\!\! &\!\!0.831\!\! & \!\!\textbf{0.879}\!\! &\!\!0.839\!\! &\!\!0.876\!\! & \!\!\textbf{0.915}\!\! &\!\!0.862\!\! &\!\!0.903\!\! & \!\!\textbf{0.934}\!\! \\
\hline
\end{tabular}
}
\label{table4}
\end{table*}

\begin{figure}
  \centering
      \includegraphics[width=1 \linewidth]{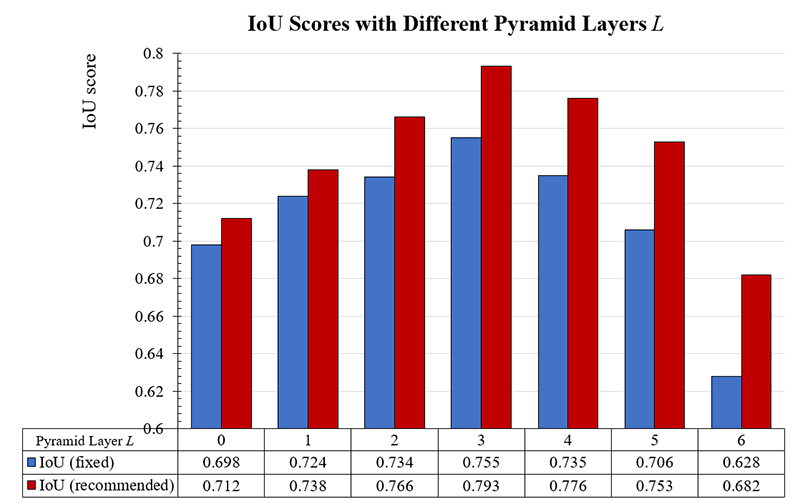}
\caption{Average IoU score with different pyramid layers $L$ on JumpCut dataset \cite{Fan2015}. The \textit{fixed} indicates the performance with fixed annotation frames (the first and the last frames), and the \textit{recommended} refers to the results using two annotation frames recommended by our method.}
  \label{fig:exp_PyramidLayer}
\end{figure}

\subsubsection{Performance of Pyramid Model}
To demonstrate the effectiveness of our global pyramid appearance model, we evaluate the performance with different pyramid resolution levels. With a foreground probability map estimated via pyramid classifier, we obtain a segmentation via a simple threshold (0.5). We choose the first 100 frames of each test video of JumpCut dataset~\cite{Fan2015} and use a pair of annotated masks of the beginning and the end frames to build the global pyramid classifiers. We set the max resolution level $L = 0, {\cdots}, 6$. Additionally, as our annotation frame selection prompts pyramid appearance model, we investigate the performance of our approach with two annotation frames recommended by our annotation frame selection system.

The overall IoU scores on JumpCut dataset~\cite{Fan2015} with different pyramid layers are presented in Fig.~\ref{fig:exp_PyramidLayer}. Note that, when the max pyramid layer $L = 0$, our pyramid classifier is equal to traditional single-level color histogram model. Overall, the performance of the pyramid classifier increases with cumulatively larger pyramid layers ($L$$\uparrow$), since, obviously, finer grid size is used and more localized structure information is considered. However, the classification performance drops as pyramid layer $L$ continues to increase. This is mainly due to the overemphasis of location information. The maximum performance is obtained when $L = 3$.

Additionally, it is clear that the classification performance with the recommended annotation frames consistently better than that with fixed annotation frames for all the pyramid layers, which confirms the annotation frame selection system plays a positive role in promoting the pyramid appearance classifier.

\subsection{User Study}
\label{sec:US}
We conducted a user study on JumpCut dataset~\cite{Fan2015} to assess the degree to which our annotation frame selection system can reduce the user effort and improve the segmentation accuracy. A corpus of 10 participants (6 female) with diverse backgrounds and ages were recruited to participate in our user study. 12 video clips, each with 100 frames, are used for training the participants with our cutout system and Rotobrush tool \cite{Bai2009}. In the training process, the participants were asked to select $K$ annotation frames ($K = 1, {\cdots}, 5$) for obtaining the best possible segmentation results. After the participants had confirmed they were familiar with the cutout systems, they were asked to segment 5 test video clips via two cutout methods. Those test video sequences are different from the training ones, and each of them also has 100 frames. The participants were presented with the test video and tried to select $K$ annotation frames for segmenting as accurately as possible. Thus a total of 50 video cutout tasks were assigned to each user. To exclude the influences of difference between annotations initialized by different users, we use the same annotations for all participants in the testing process. Our method and Rotobrush tool worked in the bi-directional workflow.

We recorded the average time that participants took to select the annotations and computed the average IoU score using the ground truth masks. To gain a deeper insight, we further offer a baseline (\textit{random}) for both two methods: generates segments via randomly selecting $K$ annotation frames. The results of the study are summarized in Table~\ref{table4}, averaged over all users for each task. It shows the human decision is positive for cutout task as the results with human participation are generally better that with randomly selected annotations. However, according to user feedbacks, scanning all the frames has already been consumed much time and effort, let alone determining which frames should be best selected. These results confirm that our annotation frame selection system successfully reduces the user effort and improves the segmentation performance. It also demonstrates that our cutout system consistently outperforms Rotobrush.

\subsection{Runtime Analysis}
\label{sec:RA}
\begin{table} [!htbp]
\centering
\caption{Comparison of average run time (seconds per frame) with $480p$ video clips of DAVIS dataset \cite{Perazzi2016}.}
\renewcommand\arraystretch{1.1}
\begin{threeparttable}
\resizebox{0.49\textwidth}{!}{
\begin{tabular}{|c||c|c|c|c|c|c|}
\hline
Method &Ours &RB09 &HV10 &SS14 &JC15 &BV16\\
\hline
Time(s) &4.23 &0.62 &5.67 &6.36 &13.4 &-~(0.37)\tnote{*}\\
\hline
\end{tabular}
}
\begin{tablenotes}
\footnotesize
\item[*] The runtime reported in BV16 is offered for reference, since the code in
{https://github.com/owang/BilateralVideoSegmentation} is much slower.
\end{tablenotes}
\end{threeparttable}
\label{table6}
\end{table}

We measure the running time of the proposed method and the current fastest video cutout and segmentation tracking methods: RB09~\cite{Bai2009}, HV10~\cite{grundmann2010}, SS14~\cite{Ramakanth2014},  JC15~\cite{Fan2015}, BV16~\cite{maerki2016}. We directly run their publicly available codes. We carried out time consumption analysis on a personal computer equipped with Intel Core 2 Duo E8400 3-GHz CPU and 4GB RAM.
All the tests were performed on the $480p$ video clips of DAVIS dataset \cite{Perazzi2016}. As shown in Table~\ref{table6}, our method is faster than the others except RB09. Considering our promising results, our method achieves
a better tradeoff between performance and computation efficiency.

\section{Conclusions}
\label{sec:conclusion}
This article presented a video cutout technique that efficiently propagates user annotations for whole video sequence.
We propose a pyramid histogram appearance classifier, which considers structure information into color statistics. It works by repeatedly subdividing an image and computing histograms of image features over the resulting subregions, has shown promising results over traditional classifiers.
The pyramid classifier is completed with a set of geodesic distance based dynamic foreground models, for jointly identifying the object. Then, local classifiers are adopted for integrating multiple local image features, such as color and shape information. The final cutout results are achieved by the collaboration of global classifiers and local classifiers working on those uncertain regions. Additionally, the proposed method intelligently selects the optimal frame for human labeling, which reduces  manual effort and improves segmentation performance.
Comprehensive evaluation and comparison results on two databases, namely DAVIS and JumpCut, have clearly demonstrated the effectiveness of the proposed approach at achieving promising results. Additionally, the extensive user study has verified the proposed intelligent  annotation frame selection greatly contributes to the segmentation efficiency and boosts final performance.

{\small
\bibliographystyle{IEEEtran}

}

\vfill


\end{document}